\documentclass{article}
\pdfoutput=1
\usepackage{microtype}
\usepackage{graphicx}
\usepackage{subfigure}
\usepackage{booktabs} % for professional tables
\usepackage{hyperref}

\usepackage{amsmath,amssymb, amsthm}
\usepackage{bm}
\usepackage{multirow} % For merging tables rows
\usepackage[accepted]{icml2021}
\usepackage[colorinlistoftodos,prependcaption,textsize=small]{todonotes}
\usepackage{xargs}
\usepackage{xstring}
\def\G{\mathcal{G}}
\def\V{\mathcal{V}}
\def\v{\mathbf{v}}
\def\x{\mathbf{x}}
\def\f{\mathbf{f}}
\def\R{\mathbb{R}}
\def\E{\mathbb{E}}
\def\D{\mathcal{D}}
\def\m{\mathbf{m}}
\def\GP{\mathcal{GP}}
\def\K{\mathbf{K}}
\def\N{\mathcal{N}}
\def\cov{\mathrm{cov}}
\icmltitlerunning{Evolving-Graph Gaussian Processes}

\begin{document}
	
	\twocolumn[
	\icmltitle{Evolving-Graph Gaussian Processes}
	
	
	
	
	\begin{icmlauthorlist}
		\icmlauthor{David Blanco-Mulero}{see}
		\icmlauthor{Markus Heinonen}{css}
		\icmlauthor{Ville Kyrki}{see}
	\end{icmlauthorlist}
	
	\icmlaffiliation{see}{School of Electrical Engineering, Aalto University, Espoo, Finland}
	\icmlaffiliation{css}{Department of Computer Science, Aalto University, Espoo, Finland}

	\icmlcorrespondingauthor{David Blanco-Mulero}{david.blancomulero@aalto.fi}
	
	\icmlkeywords{Machine Learning, ICML}
	
	\vskip 0.3in
	]
	
	\printAffiliationsAndNotice{} % leave blank if no need to mention equal contribution
	
	\begin{abstract}
		Graph Gaussian Processes (GGPs) provide a data-efficient solution on graph structured domains. 
		Existing approaches have focused on static structures, whereas many real graph data represent a dynamic structure, limiting the applications of GGPs.
		To overcome this we propose evolving-Graph Gaussian Processes (e-GGPs).
		The proposed method is capable of learning the transition function of graph vertices over time with a neighbourhood kernel to model the connectivity and interaction changes between vertices. 
		We assess the performance of our method on time-series regression problems where graphs evolve over time.
		We demonstrate the benefits of e-GGPs over static graph Gaussian Process approaches.
		%and outperform graph neural network (GNN) approaches when the training data is scarce.
	\end{abstract}

	\section{Introduction}

	\label{introduction}
	
	% 1-Introduce dynamic graphs
	Dynamic graphs provide rich representations for temporal structured data to model evolving relationships in the data.
	Structured data is often dynamic, such as human interactions \cite{casteigts_TVgraphs_human_interactions_2012}, social networks \cite{trivedi_dyrep_dynamic_graph_socialnets_2019} or brain interactions \cite{ofori-boateng_dynamic_graph_fmri_2020}.
	% Introduce here evolving graph and other terms in the literature
	These structures have been studied as dynamic graphs, time-varying structures \cite{lebars_timevarying_ising_model_2020}, and \textit{evolving graphs} \cite{xuan_evolving_graphs_2002}.
	% Problem with current methods that solve dynamic graphs
	However, the predominant methods for estimating dynamic graph evolution fail to account for model uncertainty \cite{sanchez-gonzalez_GNs_learningphysics_for_control_2018, li_DPINets_2019, zhu_dynamic_graph_learning_2019}.
	This uncertainty can be crucial in dynamic systems where safety constraints are required \cite{Hewing_Cautious_MPC_GP_2020}.
	
	% 2- Introduce GPs, mention data-efficiency as well?
	A natural choice to quantify uncertainty of a model are Gaussian processes (GPs) \cite{rasmussen_GPbook_2006}. %, which are Bayesian function learning models.
	% 3-Introduce AR-GPs and limitation, graph structured data
	Pioneering attempts of applying Gaussian processes to model dynamics estimate autoregressive temporal transition functions with GPs in the Euclidean domain \citep{wang2005gaussian,mattos2015recurrent}.
	% 4- Now move to graph-GPs
	More recently, GPs have been applied to problems on graph-structured domains (graph GPs), such as signal processing over graphs \cite{ walker_graph_2019, venkitaraman_gaussian_2020, li2020stochastic} or link prediction \cite{opolka_graph_2020}.
	% What can graph-GPs do?
	Some of these works decompose the structured domain through graph convolutions \cite{walker_graph_2019, opolka_graph_2020} but their output is limited to the Euclidean domain.
	Others, provide an output in the graph-domain \cite{venkitaraman_gaussian_2020, zhi_gaussian_2020} but are restricted to a vectorial input.
	% 4- Issue with graph-GPs
	%The main issue of these works is that they fail to account for the evolving structure of the graph.
	To our knowledge, no Gaussian process models have been proposed to model temporal evolution of graph  structures.
	
	\begin{figure}[t]
		\vskip 0.1in
		\centering
		\includegraphics[width=0.65\columnwidth]{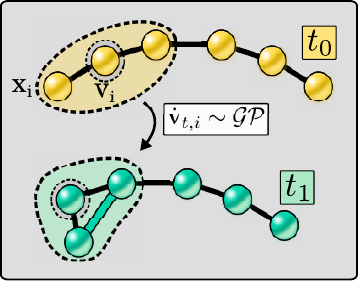}
		\caption{Evolving graph where the neighbourhood of the vertex $\mathbf{v}_i$ changes  and the transition is modelled by an evolving-Graph Gaussian Process (e-GGP).}
		\label{fig:cool-intro-evolution}

	\end{figure}
	
	We present an autoregressive GP model that learns the transition function of vertices $f : \v \mapsto \dot{\v}$ on the graph domain $G$, which we call the evolving-Graph Gaussian Process (e-GGP), see Figure~\ref{fig:cool-intro-evolution}.
	We define a graph kernel that considers the neighbourhoods and attributes of the graph.
	We investigate the capability of our method to learn the graph evolution in simulated physical dynamical systems, where a measure of the simulation uncertainty can be required.
	Our results showcase the ability of e-GGP to learn accurate graph dynamics, and extrapolate the dynamics to graphs with unseen structures at test time.

	\section{Background}
	\label{sec:background}

	\paragraph{Graph-based Gaussian processes.}
	There is a rich literature of different configurations of Gaussian processes over graph domains. 
	As previously mentioned, we can distinguish between two configurations. The first, maps from a vectorial input to an output in the graph domain $f : \R^D \rightarrow G$ \citep{venkitaraman_gaussian_2020,zhi_gaussian_2020}. The second, takes an input graph and maps it to the Euclidean domain $f : G \rightarrow \R$ \citep{ng_bayesian_2018, walker_graph_2019, opolka_graph_2020, borovitskiy_matern_2020}. For a more thorough overview of graph GP methods see Supplementary Material Section ~\ref{supp:tab-overview-gps}.
	These methods consider a fixed adjacency matrix of the graph input, or use convolution operations \citep{van_der_wilk_convolutional_2017}, which narrows their input to a static graph structure.

	\paragraph{Gaussian processes for time-series.}
	The Gaussian process dynamical model (GPDM) models autoregressive Markov transitions of discrete-time random states with a Gaussian process \citep{wang2005gaussian}. In latent modelling a Gaussian process interpolates the object trajectories in latent space \citep{damianou2011variational}. More generally these models fall under recurrent Gaussian processes \citep{mattos2015recurrent}. The GPAR extends multi-output Gaussian processes to time-series \citep{requeima19}. These models are limited to vector-valued inputs, and do not support non-Euclidean inputs. 
	
	The most relevant prior work to our aims is the deep Graph Gaussian process (DGPG) \citep{li2020stochastic}, where the graph structure is auto-regressed before applying a conventional classification or regression step. However, their approach is limited to a fixed graph structure.

	\section{Evolving Graphs using Gaussian Processes}
	\label{sec:method-evolving-graphs}

	We consider a temporally evolving, or \emph{dynamic}, undirected graph $G_t = \langle \V_t, \mathcal{E}_t \rangle$, where $\V_t$ is the set of vertices and $\mathcal{E}_t \subset \V_t \times \V_t$ is the set of edges at discrete timepoints $t \in N$ of the graph time-series $G=\{G_0, \ldots, G_N\}$.
	For notation simplicity we consider the number of vertices in each graph to be the same $\lvert \V_t \rvert=M$.

	The vertices $\v_t \in \mathcal{V}$ reside in a $D$-dimensional space. For our objective of learning physics simulations, a vertex at time $t$ is represented as 
	$\v_t := (\x_t, \Delta \x_{t-1}, \phi(\v_t))$
	over the Cartesian coordinates $\x$, and static attributes $\boldsymbol{\phi}(\v)$, which describe node properties or types. 
	We assume that edges are a function of vertices and that the vertices-to-edges function $\mathbf{g}: 2^\V \rightarrow \mathcal{E}$ is defined by a distance function $d(\v_i, \v_j)$. 
	We adopt the L2-norm distance between the vertices Cartesian coordinates. Therefore, an edge exists if the distance between two nodes is lower than a connectivity threshold $R_{nn}$, see Supplementary Material Section~\ref{supp:main-training}.
	
	We assume the graph dynamics are governed by
	\begin{align}
	\label{eq:graph-markov-dynamics}
	G_{t+1} = T(G_t)
	\end{align}
	driven by the graph transition function $T:G \rightarrow G$, 
	where the graph transition is defined by a transition function of node attributes $\V_{t+1} = T(\V_t)$ and the edges are given by the nodes-to-edge function.
	% starting from an initial graph state $G_0$. 
	% Define graph summation = node summation + edges given by function g()
	Our goal is to learn the evolution function $T$.
	Hence, we opt to model and learn the node evolution in the discrete timestep $\Delta t$ by:
	\begin{align}
	\label{label:dyn1}
	\v_{t+1} &= \v_t + \underbrace{\f(N^{0,1}(\v_t))}_{\Delta \v_t}, 
	\end{align}
	with a velocity function $\f : s(G) \rightarrow \R^F$ mapping a sub-tree $s(G)$ into coordinate velocities $\R^F$. The sub-tree $N^{0,1}(\v_t)$ consists of a node $\v_t$, or $N^0(\v_t)$, and its neighbours $N^1(\v_t)$.
	To simplify the notation in the following, we will use $\f(\v_t)$ to denote the mapping. 
	This formulation allows us to handle node insertion and deletion.

	\subsection{Prior and likelihood}
	
	We model the velocity transitions with a vector-valued Gaussian process (GP) prior \citep{rasmussen_GPbook_2006,alvarez2011}
	\begin{align}
	\label{eq:prior}
	\f(\v) &\sim \GP(\m(\v), K(N^{0,1}(\v),N^{0,1}(\v'))),
	\end{align}
	which implies that the transition is a stochastic process, whose mean and covariance are parameterised by the mean function $\m(\cdot)$ and the kernel similarity $K(\cdot,\cdot)$ as
	\begin{align}
	\E[ \f(\v) ] &= \m(\v), \\
	\cov[ \f(\v), \f(\v')] &= K(N^{0,1}(\v),N^{0,1}(\v')).
	\end{align}
	For tractability, we assume different velocity predictions are independent, however sharing the same kernel, which simplifies the kernel into a scalar function. Furthermore, for any subset of nodes $\v_1, \ldots, \v_M$ the corresponding transitions are jointly Gaussian
	\begin{align}
	p(\Delta \v_1, \ldots, \Delta\v_M) &= \N( \Delta \vec{\v} | \vec{\m}, \K),
	\end{align}
	where $\Delta \vec{\v} \in \R^{M F}$ and $\vec{\m} \in \R^{M F}$ stack all vertices and means to column vectors, and $\K \in \R^{MF \times M F}$ is the block kernel matrix consisting of $M \times M$ blocks $K(\v_i,\v_j)$ of size $F \times F$. The key property of Gaussian processes is that similar nodes have similar transitions, as specified by the similarity function $K$.

	We assume a dataset
	\begin{align}
	\mathcal{D} &= \Big\{ (\v_{t,i}, \Delta \v_{t,i}) \Big\}_{t=1,i=1}^{N,M}    
	\end{align}
	of observed graph states $G_t$ that are described in terms of the $M$ vertices at $N$ timepoints. Our observation model  induces a likelihood
	\begin{align}
	\label{eq:lik}
	p(\D | \f) &= \prod_{t=1}^N \N\left( \v_{t,i} | \f, \Sigma\right),
	\end{align}
	where $\f$ is the estimated node evolution following equation \eqref{label:dyn1}, and $\Sigma \in \R^{F \times F}$ represents the numerical variance required for numerical stability.
	
	\subsection{Posterior}
	
	The predictive posterior distribution of our Gaussian process model is conveniently tractable under the Gaussian likelihood model \eqref{eq:lik} as \citep{rasmussen_GPbook_2006}
	\begin{align}
	p(\f_* | \D) &= \N(\f_* | \boldsymbol{\mu}_*, \Sigma_*), \\ 
	\boldsymbol{\mu}_* &= \K_* (\K + \mathbf{\Sigma})^{-1} \Delta\vec{\v}, \\
	\Sigma_* &= \K_{**} - \K_* (\K + \mathbf{\Sigma})^{-1} \K_*^T,
	\end{align}
	% Dimension N graphs |V|=M vertices
	where $\K \in \R^{NMF \times NMF}$, $\K_* \in \R^{N_*MF \times NMF}$ and $\K_{**}\in \R^{N_*MF \times N_*MF}$ for a test set consisting of $N_*$ graphs with $M$ nodes. The posterior effectively interpolates the velocity predictions from the observed velocities of the training graphs.

	\subsection{Kernel between attributed sub-trees}
	We now define a kernel that can measure the similarity between vertices and their transitions.
	We propose to use a kernel on the graph that measures the similarity between attributed sub-trees $s(G)$, $K: s(G) \times s(G) \rightarrow \R$. For simplicity, we refer to the kernel as $K(\v_i, \v_j)$.
	The kernel requires the attributed vertices to be in the same space.
	Therefore, we define a vertex mapping function $\varphi_\v$ which maps the attributed vertex to an Euclidean-space $\varphi_\v: \R^D \mapsto \R^E$. In our case, $\varphi_\v$ discards the static attributes of the vertices.

	The kernel input sub-trees are rooted at the vertices $\v_i$ and $\v_j$ with 1-hop neighbourhood.
	% Roots
	We measure the similarity between the attributed roots using the \textit{root kernel} $k_r: \R^E \times \R^E \rightarrow \R$.
	% Neighbours
	In order to share information with the root and compare the structures, we compute a kernel between all the attributed leaves in the neighbourhood $N_i^1 = N^1(\v_i)$. The similarity between the leaves is measured by the \textit{leaf kernel} $k_l: \R^E \times \R^E \rightarrow \R$.
	% Mapping
	Both the \textit{root} and the \textit{leaf} \textit{kernel} operations are performed in the mapped space $\R^E$.
	% Neighbourhood
	We express the neighbourhood kernel $k_{nn}(\v_i, \v_j)$ as
	\begin{equation}
	k_{nn}(\v_i, \v_j) = \dfrac{1}{L} \sum\limits_{\x_i \in N_i^1}
	\sum\limits_{\x_j \in N^1_j} k_l(\varphi_\v(\x_{i}), \varphi_\v(\x_{j})),
	\end{equation}
	where $L$ is a normalisation constant that denotes the product between the number of leaves in $N^1_i$ and $N^1_j$.
	% Full kernel equation
	Then, the kernel between two attributed vertices is defined by
	\begin{equation}
	\label{eq:kernel-eq}
	K(\v_i,\v_j) = k_r(\varphi_\v(\v_{i}), \varphi_\v(\v_{j})) + k_{nn}(\v_i, \v_j).
	\end{equation}
	
	We restrict the kernel to the 1-hop neighbourhood for scalability and plan to investigate the effect of the neighbourhood in future work.

	%%%%%%%%%%%%%%%%%%%% EXPERIMENTS %%%%%%%%%%%%%%%%%%%%%%
	\section{Experimental results}
	\label{sec:experiments}
	
	\subsection{Experiments description}
\begin{figure}[t]
		\vskip 0.1in
		\centering
		\subfigure[]{\label{fig:rope_sphere-a}\includegraphics[width=0.48\columnwidth]{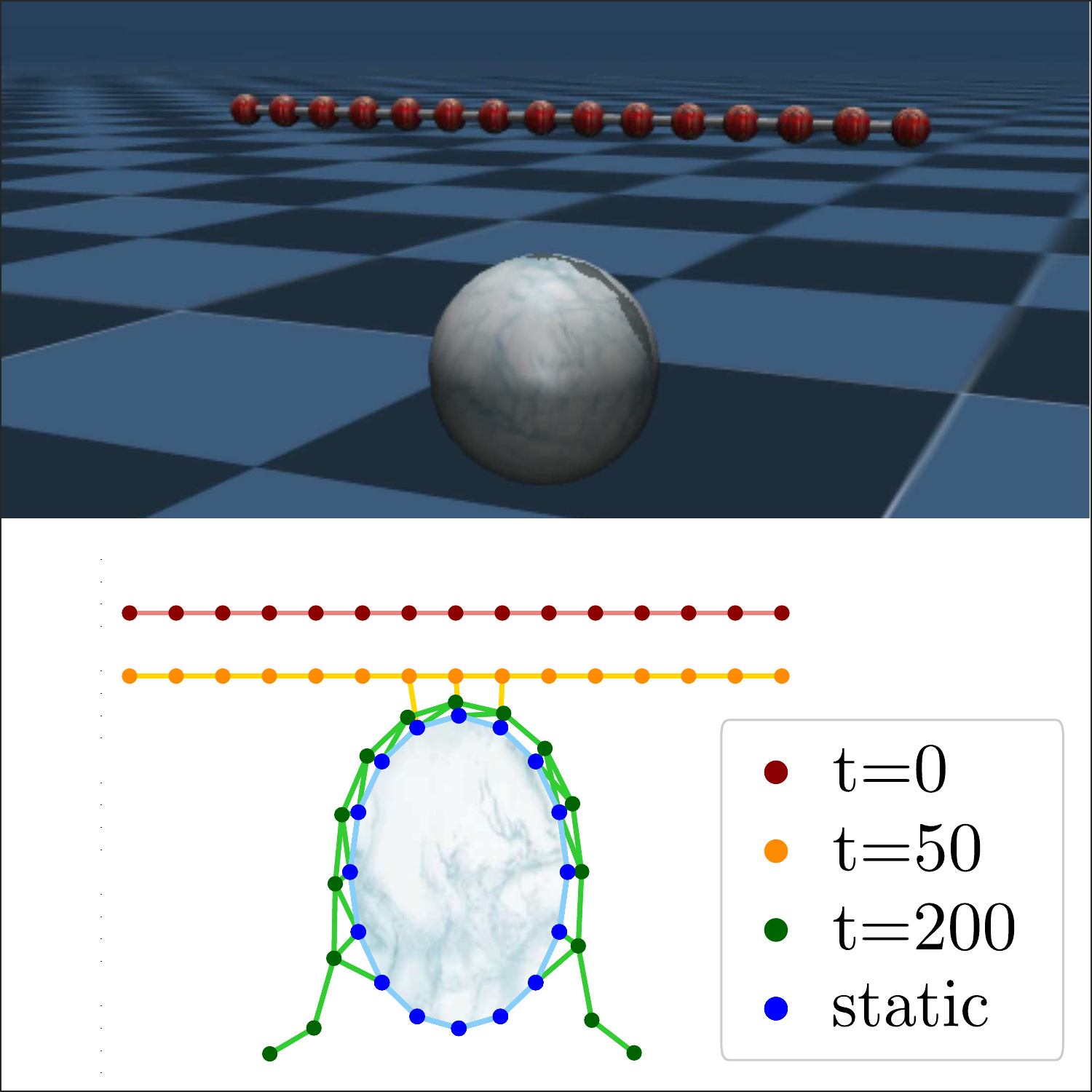}}
		\subfigure[]{\label{fig:water-2d-b}\includegraphics[width=0.48\columnwidth]{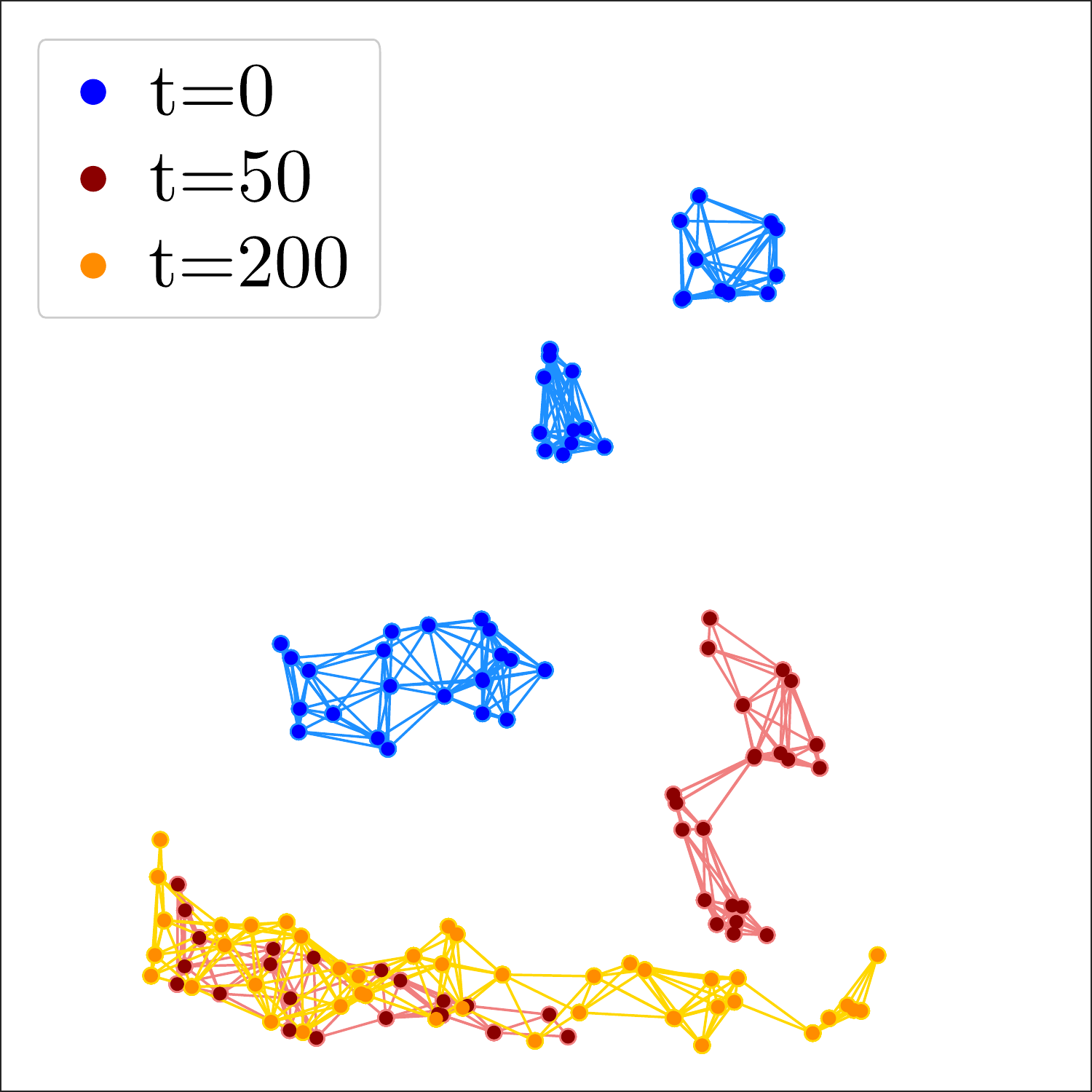}}
		\caption{
			Physics simulations: (a) graph interaction environment in MuJoCo simulation and 2-D approximation,
			(b) evolving isolated sub-graphs environment, showing graphs connectivity for different timepoints $t$.}
		\label{fig:environments}
		\vskip 0.1in
	\end{figure}

	% Intro to experiments and objective
	% 1) evaluate the importance of evolving graphs
	% 2) compare the performance of e-GGP against other GP methods.
	We evaluate the performance of e-GGPs to learn the node transition function $\Delta \v_t=\f(\v_t)$ in two physics simulations where graphs evolve over time: 1) graph interaction (GI) environment, and 2) evolving isolated sub-graphs (EIs) environment.
	The GI environment simulates a rope on free fall that gets in contact with a static ball using MuJoCo \cite{Todorov_MuJoCo_2012}.
	The EIs environment simulates a set of 2D fluid particles using Taichi \cite{Yuanming_TaichiMPM_2018}.
	Both simulations are represented in 2-D, where the joints of the rope, the static ball and the particles define the set of nodes $\mathbf{p}_t=\v_t\in\V$, see Figure~\ref{fig:environments}.
	Thus, the experiments consist of multiple sub-graphs where connectivity varies over time. 
	
	% Our environments simulate the time-series $G = \{G_0, \dots, G_N\}$.
	The purpose our experiments is twofold: (i) investigate the capability of the methods to address the varying connectivity and (ii) validate that e-GGP can learn the evolution of the graph by capturing the interaction changes between vertices.
	
	\begin{table*}
		\vskip 0.1in
		\caption{Average results of GPR, GPAR, DGPG and e-GGP on test sets (lower is better) for different training data sizes $N$.}
		\label{tab:exp-comparison-rope-ball}
		\begin{center}
			% \resizebox{\linewidth}{!}{
			\begin{tabular}{c c c c c c  c c c c}
				\toprule
				& & \multicolumn{4}{c}{Graph interaction} & \multicolumn{4}{c}{Evolving isolated sub-graphs} \\
				\cmidrule(lr){3-6}  \cmidrule(lr){7-10}
				$N$ & \textbf{Metric} & \textbf{GPR} & \textbf{GPAR} & \textbf{DGPG} & \textbf{e-GGP}
				& \textbf{GPR} & \textbf{GPAR} & \textbf{DGPG} & \textbf{e-GGP} \\
				\toprule
				\multirow{3}{*}{10}
				& RMSE    &	0.121    &	0.143    &	0.413    &	\textbf{0.049}
				&	0.294 &	0.283 &	0.337 &	\textbf{0.182}\\
				& MAPE    &	0.418    &	0.537    &	2.456    &	\textbf{0.249}
				&	0.873 &\textbf{0.726} &	2.696 &	\underline{0.777}\\
				& NLL    &	67.63    &	\textbf{-9.16}    &	9.51    &	10.44
				&	43.37 &	-47.942 &	22.24 &	\textbf{-58.042}\\
				\midrule
				\multirow{3}{*}{15}
				& RMSE    &	0.149    &	0.209    &	0.275    &	\textbf{0.031}
				&	0.276 &	0.277 &	0.325 &	\textbf{0.183}\\
				& MAPE    &	0.410    &	0.397    &	1.873    &	\textbf{0.230}
				&	\textbf{0.671} &	\textbf{0.671} &	3.135 &	\underline{0.853}\\
				& NLL    &	102.49    &	-7.67    &	2.20    &	\textbf{-17.56}
				&	103.82 &	-46.764 &	20.12 &	\textbf{-78.58}\\
				\midrule
				\multirow{3}{*}{20}
				& RMSE    &	0.162    &	0.250    &	0.222    &	\textbf{0.036}
				&	0.257 &	0.253 &	0.300 &	\textbf{0.160}\\
				& MAPE    &	0.426    &	0.466    &	1.455    &	\textbf{0.148}
				&	\underline{0.612} &	\textbf{0.594} &	2.998 &	0.766\\
				& NLL    &	126.97    &	-6.84    &	-2.18    &	\textbf{-29.69}
				&	4.17E6 &	\textbf{-48.52} &	9.731 &	8.77E5\\
				\bottomrule
			\end{tabular}
			% }
		\end{center}
	\end{table*}
	
	\subsection{Baselines and training details}
	\label{sec:exp-eval-details}
	We consider three Gaussian process methods as baselines: GP regression (GPR) using exact inference, GPAR and DGPG.
	The first two baselines are GP functions that model transition functions in the Euclidean domain.
	We use baselines in the Euclidean domain to demonstrate the potential of graph-structured information compared to approaches restricted to a vector-valued input.
	For these we take the vertices attributes vector $\V$ as input.
	Then, we learn a mapping $\f: \R^{D} \rightarrow \R^F$.
	The mapping of DGPG is defined as $\f:G \rightarrow \R^F$.
	Since DGPG does not handle node insertions we assume the adjacency matrix to be equal to the initial state graph adjacency for any given timepoint $A_t=A_0$.

	The kernel hyper-parameters are trained using automatic relevance determination (ARD) and optimised by minimising the negative marginal log-likelihood (MLL) \cite{rasmussen_GPbook_2006}.
	For all the methods we used as optimizer Adam \cite{Diederik_Adam_2015}.
	We use a Radial Basis Function kernel for the \textit{root} and \textit{leaf} kernels $k_r$ and $k_l$ in e-GGP and GPR. In DGGP we use the Matérn32 kernel and $Z=5$ inducing points.
	% Both e-GGP and GPR are implemented on GPyTorch \cite{Gardner_GPyTorch_2018}.
	% For GPAR we use the standard implementation given by the authors\footnote{https://github.com/wesselb/gpar} and learn a single output at a time (we do not learn dependencies between the outputs).
	% For GPAR we learn a signle output at a time (we do not learn dependencies between the outputs). 
	% For GPAR we use the standard implementation given by the authors\footnote{https://github.com/wesselb/gpar} and learn a single output at a time (we do not learn dependencies between the outputs).
	% In the case of DGPG, we use the authors' implementation\footnote{https://github.com/naiqili/DGPG} with $M=5$ inducing points per node and a single layer with the Matérn32 (M32) kernel.
	For more details about the training see Supplementary Material Section~\ref{supp:main-training}.

	\subsection{Results}
	\label{sec:exp-scarce-data}
	We investigate the error of the node evolution predictions.
	In order to evaluate the methods ability to capture the evolving graph information we limit the amount of data provided in training. We select the most informative training points as described in Supplementary Material Section~\ref{supp:main-training}.
	To evaluate the prediction performance and confidence we measure the root mean squared error (RMSE), mean absolute percentage error (MAPE) and negative log-likelihood (NLL).
	The test datasets on the GI environment include an offset between the rope and the static ball from the training data to evaluate the generalisation capabilities of the methods.
	We provide additional results in Supplementary Material Section~\ref{supp:main-results}.
	
	The results for the GI environment on Table~\ref{tab:exp-comparison-rope-ball} show that e-GGP outperforms the baselines when data is scarce.
	The different orders of magnitude in the RMSE provide evidence that the method we propose benefits of the evolving graph-structured information captured by the kernel.
	These results also provide evidence that e-GGP can generalise better to unseen data under limited training samples.
	Furthermore, e-GGP presents low RMSE for the EIs environment, which shows that the model can generalise to unseen sub-graph structures, i.e. sub-graphs that have not been seen during training.
	When increasing the sample size to $N=20$ the confidence drops while keeping low RMSE and MAPE.
	This suggests that the model is not able to provide a confident estimate under sparse data, which will be investigated in the future.
	The DGPG method is limited to a fixed graph size and cannot handle node insertions, which is shown by the higher error results.

	In summary, the results provide evidence that: (i) e-GGP can predict the node evolution under varying connectivity, and (ii) learning the interactions between the evolving vertices is beneficial as e-GGP outperforms the baselines in the limited data setting.

	\section{Conclusion}
	\label{conclusion}
	
	In this paper we proposed evolving-Graph Gaussian Processes (e-GGPs), an autoregressive graph-GP model that learns the evolving structure of dynamic graphs via the node transitions.
	Our model is able to learn the node transitions as well as the interactions of the structure via an attributed sub-tree kernel that incorporates information from each node's neighbourhood.
	The experimental results demonstrate that our model is able to capture evolving graph connectivity, overcoming this limitation of current methods.
	We provide evidence of the importance of capturing the dynamic connectivity and investigated the performance of our method in scarce data.
	Our experiments showed that e-GGP outperforms other approaches by making use of the rich information in the evolving graph-domain.

	Our model opens GPs to dynamic graphs which is of great interest in real graph problems. However, our model suffers from poor scalability, which we plan to investigate via variational inference. 
	We also plan to study the application of e-GGP to real problems that require both data-efficiency and uncertainty, such as dynamic systems with safety constraints, where other approaches such as neural networks are limited.

	\section*{Acknowledgements}
	This work was supported by the Academy of Finland, decision 317020.
	
\bibliography{DynamicGraphs, Experiments, GCNs, GPs, GraphGPs, GraphKernels, GraphTheory, Safety}
\bibliographystyle{icml2021}

\newpage

\renewcommand\thefigure{\thesection.\arabic{figure}}    
\renewcommand{\thetable}{\thesection.\arabic{table}}

\setcounter{figure}{0}    
\setcounter{table}{0}

\onecolumn

\icmltitle{Supplementary Material: Evolving Graph Gaussian Processes}
\appendix

\section{Analysis of e-GGP}
\subsection{Overview of graph Gaussian processes}
\label{sec:supp-graph-gps-overview}

There is a rich literature of different configurations of Gaussian processes over graph domains. Graph-output Gaussian processes consider functions $f : \R^D \rightarrow G$, where the $|\V|$ outputs have dependencies according to an underlying output graph \citep{venkitaraman_gaussian_2020,zhi_gaussian_2020}. In semi-supervised Gaussian processes a graph relationship between inputs is assumed to propagate label information within the graph from labelled to unlabelled inputs \citep{ng_bayesian_2018}. \citet{borovitskiy_matern_2020} adapts Matérn Gaussian processes for graph node labelling. The graph convolutional Gaussian processes consider functions $f : G \rightarrow \R$, where the label of an entire input graph $G$ is predicted \citep{walker_graph_2019,opolka_graph_2020} with convolution operations \citep{van_der_wilk_convolutional_2017}.

Our method learns the transition function of vertices $f: \v \mapsto \dot{\v}$ and infers the new edges using a nodes-to-edge function $\mathbf{g}: 2^\V \rightarrow \mathcal{E}$, learning the graph transition.
We provide an overview of the different graph-GP methods as well as their mapping function to highlight the scope of our work on Table~\ref{supp:tab-overview-gps}.

\begin{table*}[h]
	\vskip 0.1in
	\caption{Overview of different graph-GP methods, their mapping functions $\f$ and whether an static graph is assumed or not. We denote each model by the name addressed in the paper or by the formulation they propose.}
	\label{supp:tab-overview-gps}
	\begin{center}
		\resizebox{\linewidth}{!}{
			\begin{tabular}{lcccccccr}
				\toprule
				& \multicolumn{3}{c}{Input domain} & \multicolumn{3}{c}{Output domain} & & \\
				\cmidrule(lr){2-4} \cmidrule(lr){5-7}
				Model 
				& $\R^D$ & $\V$ & $\G$ & $\R^D$ & $\mathcal{Y}$ & $\G$ &  Time-series & Reference \\
				\midrule
				GPDM & \checkmark & -  & - & \checkmark & - & - &  \checkmark & \citet{wang2005gaussian} \\
				GGP & - & \checkmark  & - & - & \checkmark & - &  - & \citet{ng_bayesian_2018} \\
				GCGP & -  & - & \checkmark & - & \checkmark & - &  - & \citet{walker_graph_2019,opolka_graph_2020} \\
				DGPG & -  & - & \checkmark & \checkmark & - & - & - & \citet{li2020stochastic} \\
				Matérn Graph GP & -  & - & \checkmark & \checkmark & - & - & -  & \citet{borovitskiy_matern_2020} \\
				GPG & \checkmark & -  & - & - & - & \checkmark & -  & \citet{venkitaraman_gaussian_2020} \\
				Graph-output GP & \checkmark  & - & - & - & - & \checkmark &  - & \citet{zhi_gaussian_2020} \\
				\midrule
				Ours & -  & - & \checkmark & - & - & \checkmark  & \checkmark &  \\
				\bottomrule
			\end{tabular}
		}
	\end{center}
	\vskip 0.1in
\end{table*}

\setcounter{table}{0}

\section{Supplementary training details}
\label{supp:main-training}
\subsection{Experiments data sets}
The number of nodes in each of the graph data sets is provided in Table~\ref{tab:supp-train-test-data}.
The connectivity hyper-parameter $R_{nn}$ and the maximum number of neighbours $K_{nn}$ are selected based on the environment definition.

\begin{table}[h]
	\vskip 0.15in
	\caption{Description of train and test data for the experiments described in Section~\ref{sec:experiments}. $N$ denotes the horizon, $\lvert \V \rvert = M$ the number of nodes in the graph and $D$ the attributes of the nodes.}
	\label{tab:supp-train-test-data}
	\begin{center}
		\begin{small}
			\begin{sc}
				\begin{tabular}{cccccccc}
					\toprule
					%  & & \multicolumn{2}{c}{Train data}  & \multicolumn{2}{c}{Test data} \\
					% \cmidrule{3-4}  \cmidrule{4-5}
					& N & $R_{nn}$ & $K_{nn}$ & $M_\text{train}$  & $D_\text{train}$ 
					& $M_\text{test}$  & $D_\text{test}$ \\
					\toprule
					% \midrule
					Graph Interaction           & 500 & 0.043 & 2 & 31 & 7 & 31 & 7 \\
					Isolated Evolving Sub-graphs & 200 & 0.08 & 20 & 44 & 8 & 44, 55, 66 & 8 \\
					\bottomrule
				\end{tabular}
			\end{sc}
		\end{small}
	\end{center}
	\vskip -0.1in
\end{table}

\paragraph{Graph interaction connectivity.}
We define two objects: an elastic rope and a static ball. The distance between the nodes of each object was set to $d(\v_i, \v_j)=0.03$. However, because of the elasticity of the material in MuJoCo \cite{Todorov_MuJoCo_2012}, we found out experimentally that using a connectivity lower than $R_{nn}=0.043$ would lead to disconnected nodes in the graph that are physically connected in the simulation.
Considering that each of the objects is internally connected, we use that information to build the graph connectivity. Therefore, we considered the maximum number of neighbours affecting a node to be $K_{nn}=2$, as the internal connectivity was already taken into account.

The attributes of the nodes are set to
$\v_t = \{ x_t, y_t, z_t, \Delta x_{t-1}, \Delta y_{t-1}, \Delta z_{t-1}, \mathbf{\phi}(\v_t) \}$, where $\mathbf{\phi}(\v_t)=0$ if the node belonged to the ball and $\mathbf{\phi}(\v_t)=1$ if it was part of the rope.

\begin{figure}[h]
	\vskip 0.1in
	\centering
	% 732 = 44 particles (training data)
	\subfigure[]{\includegraphics[width=0.21\columnwidth]{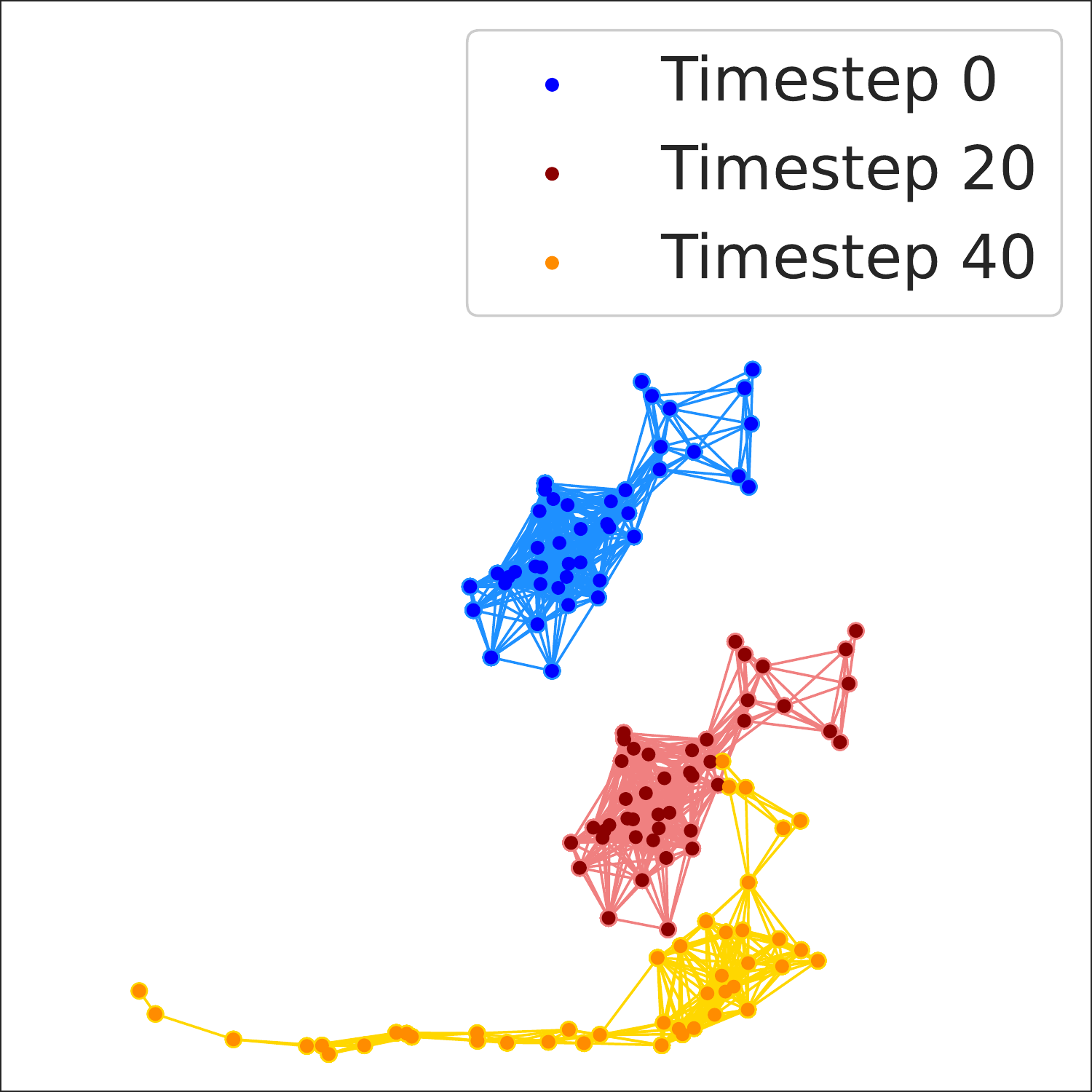}}
	% 15732 = 44 particles
	\subfigure[]{\includegraphics[width=0.21\columnwidth]{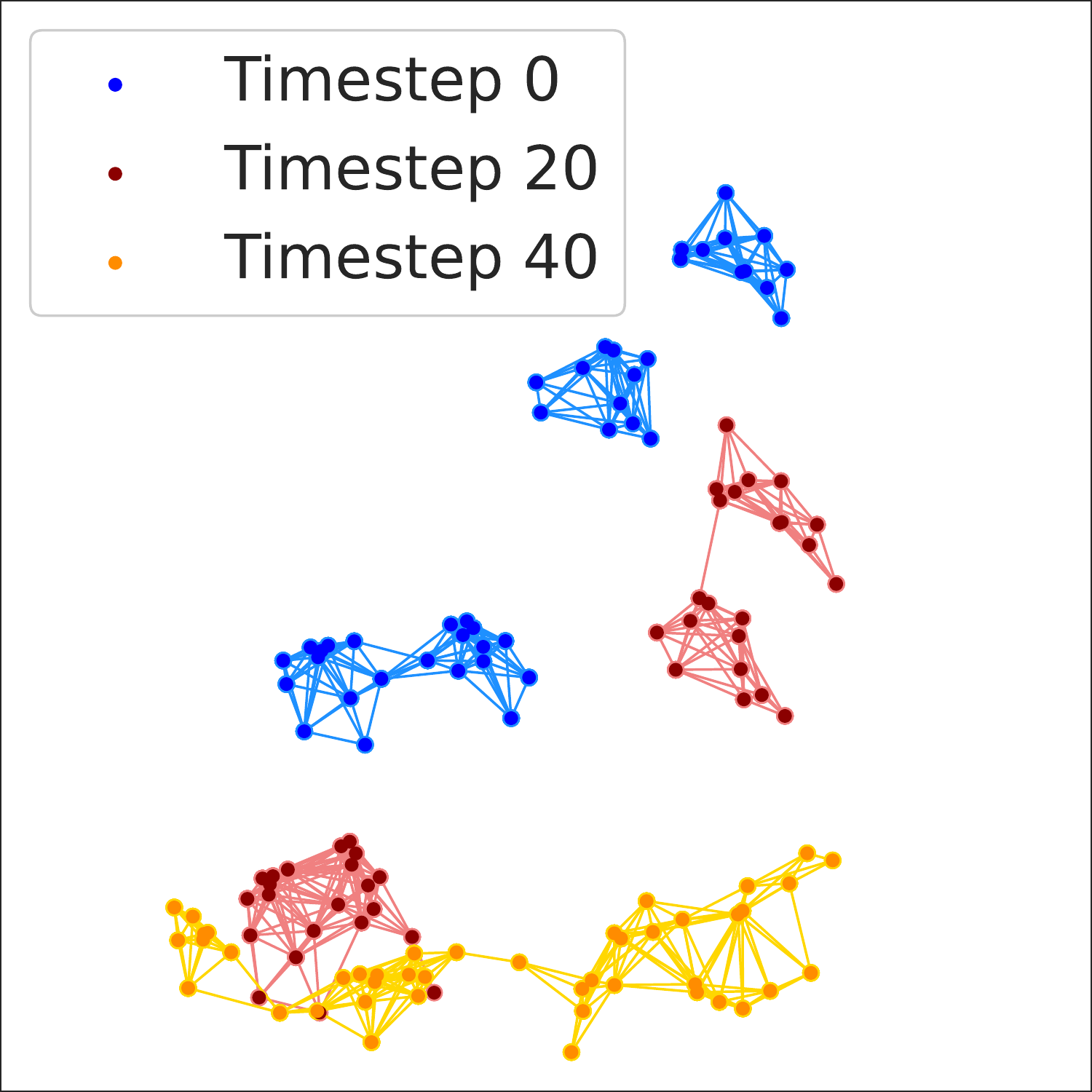}}
	% 1442 = 55 particles
	\subfigure[]{\includegraphics[width=0.21\columnwidth]{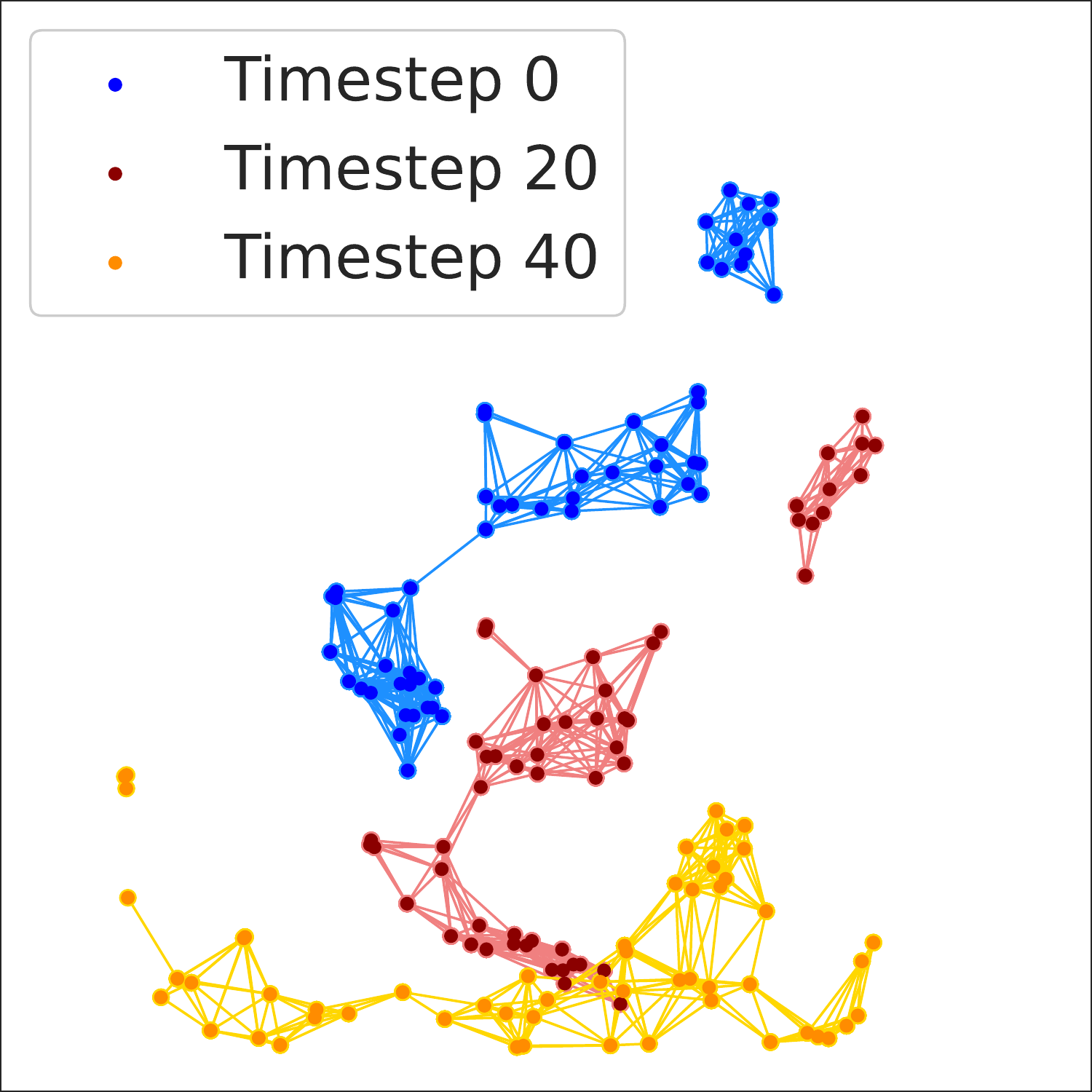}}
	% 123 = 66 particles
	\subfigure[]{\includegraphics[width=0.21\columnwidth]{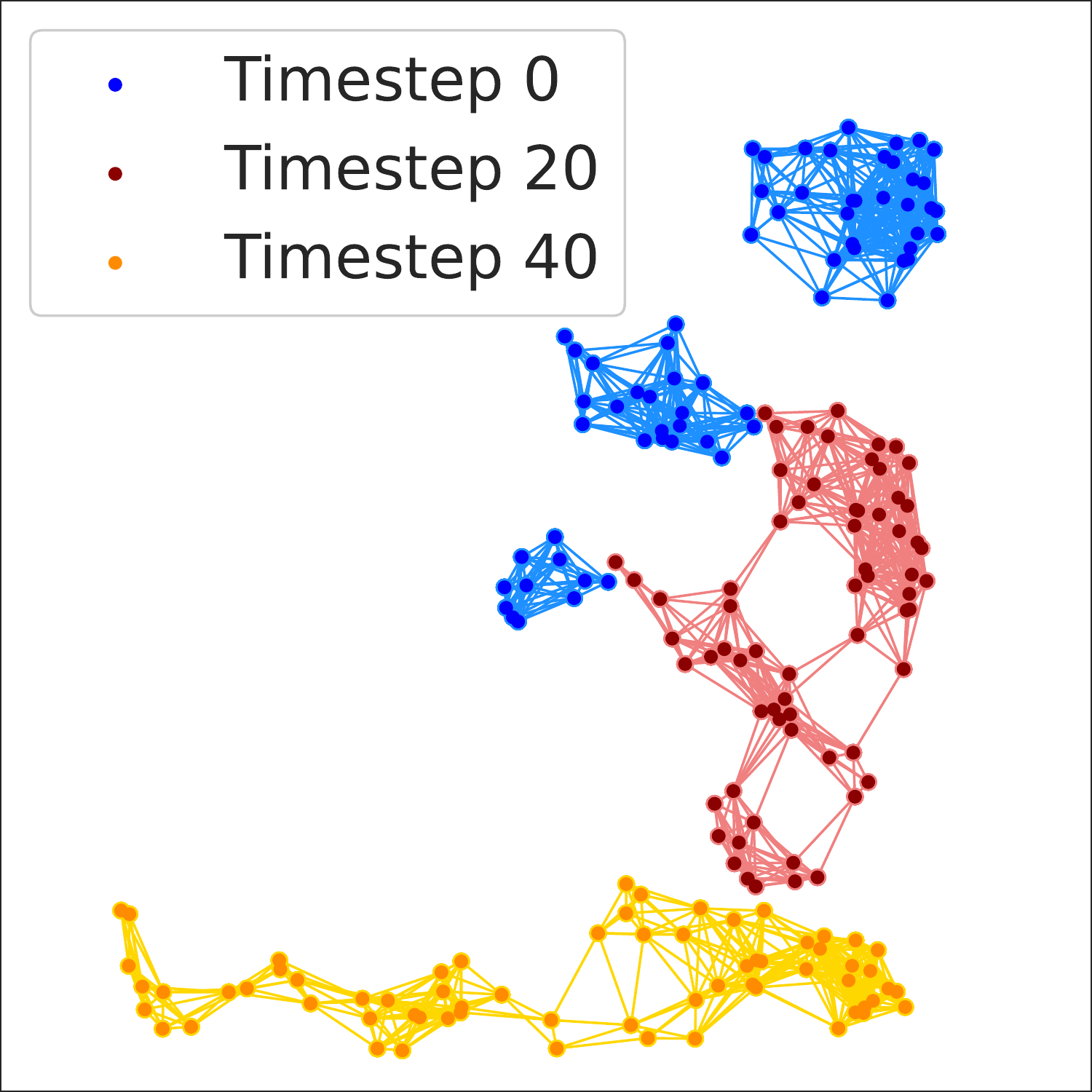}}
	\caption{Visualisation of the evolving isolated sub-graphs initial configuration, evolution and connectivity for (a) training
		data $N=44$, (b) test data $N=44$, (c) test data $N=55$, (d) test data $N=66$.}
	\label{fig:supp-particles-env}
	\vskip 0.1in
\end{figure}

\paragraph{Isolated evolving sub-graphs connectivity.}
We randomise for each test the initial conditions of the particles \cite{Yuanming_TaichiMPM_2018}. For each of the tests, we initialise a random number of blocks of particles, where the initial position and velocity are also randomised. This lead to a different evolution of the connectivity, as shown in Figure~\ref{fig:supp-particles-env}.

The resolution used in our Taich-MPM environment was set to $16$. We found out experimentally that particles that are separated by a distance of $R_{nn}=0.08$ or lower had an effect on each other so as to be considered neighbours. We decided to use $K_{nn}=20$ to provide the maximum information from the nearby particles.

In this environment, the nodes are defined as
$\v_t = \{ x_t, y_t, \Delta x_{t-1}, \Delta y_{t-1}, d(x_t, x_\text{max}), d(y_t, y_\text{max}), d(x_t, x_\text{min}), d(y_t, y_\text{min}) \}$, where $x_\text{max}$, $x_\text{min}$, $y_\text{max}$, $y_\text{min}$ refer to the boundaries of the water container.

\subsection{Kernel selection}
\label{supp:training-kernel}
The attributes of the nodes live in the coordinate space.
Therefore, we decided to use the Radial Basis Function (RBF) stationary kernel as the \textit{roof} and \textit{leaf} kernels for e-GGP as well as for the baseline of GPR.
We note to the reader that different combinations of kernels are possible when using e-GGP.
As an example, one could choose to use an RBF \textit{root} kernel and a Matérn 52 \textit{leaf} kernel.
The decision of the kernel to use for the root nodes and the neighbourhood should be based on the expert knowledge.

For the \textit{graph interaction environment}, even though the environment is approximated in 2-D, the nodes are attributed with 3-dimensional coordinates and velocities. The models are not affected by the third dimension as the kernel hyper-parameters will reject the non-informative dimensions as training is done using automatic relevance determination (ARD) \cite{rasmussen_GPbook_2006}.

\subsection{Implementation details}
\label{supp:training-implementation}
% TODO Add this only for the submission on ICML 

\paragraph{e-GGP implementation.}
We provide an open-source implementation of e-GGP\footnote{https://github.com/dblanm/evolving-ggp}.
Our implementation is based on GPyTorch \cite{Gardner_GPyTorch_2018}. 
The implementation of the \textit{root} and \textit{leaf} kernels was done so that the stationary kernels distance was scaled by a length-scale parameter $\theta_d$ per dimension $d\in D$, and a noise parameter $\sigma^2$.

\paragraph{Graph definition.}
The graph is built as follows. We take the attributed nodes at a timepoint $\V_t$ and build a k-d tree. We define the neighbourhood in the tree by the nearest K neighbours $K_{nn}$, with distance lower than the threshold $R_{nn}$, to the queried node. 
The $K_{nn}$ parameter can also be seen as the maximum degree of each node. 
The neighbours denoted by the parameters $K_{nn}$ and $R_{nn}$ define the edges of the graph $\mathcal{E}_t$, thereby forming the graph $G_t=\langle \V_t, \mathcal{E}_t \rangle$ at the given timepoint $t$. 

\paragraph{Training points selection.}
We select points from the training data set based on the rate of change of the target. Therefore, if the target is the velocity $\Delta x_t$, we select the points with highest acceleration $\Delta^2 x_t$.
In order to avoid points that are close in the temporal dimension, we restrict the time between samples to be at least $\Delta t=20$ timepoints far from each other.
In case that there are no more data points fitting the temporal constraint, the time difference is constantly reduced until the requested number of training points is satisfied.

\begin{figure}[h]
	\vskip 0.1in
	\centering
	% Coordinate spce
	\subfigure[]{\includegraphics[width=0.24\columnwidth]{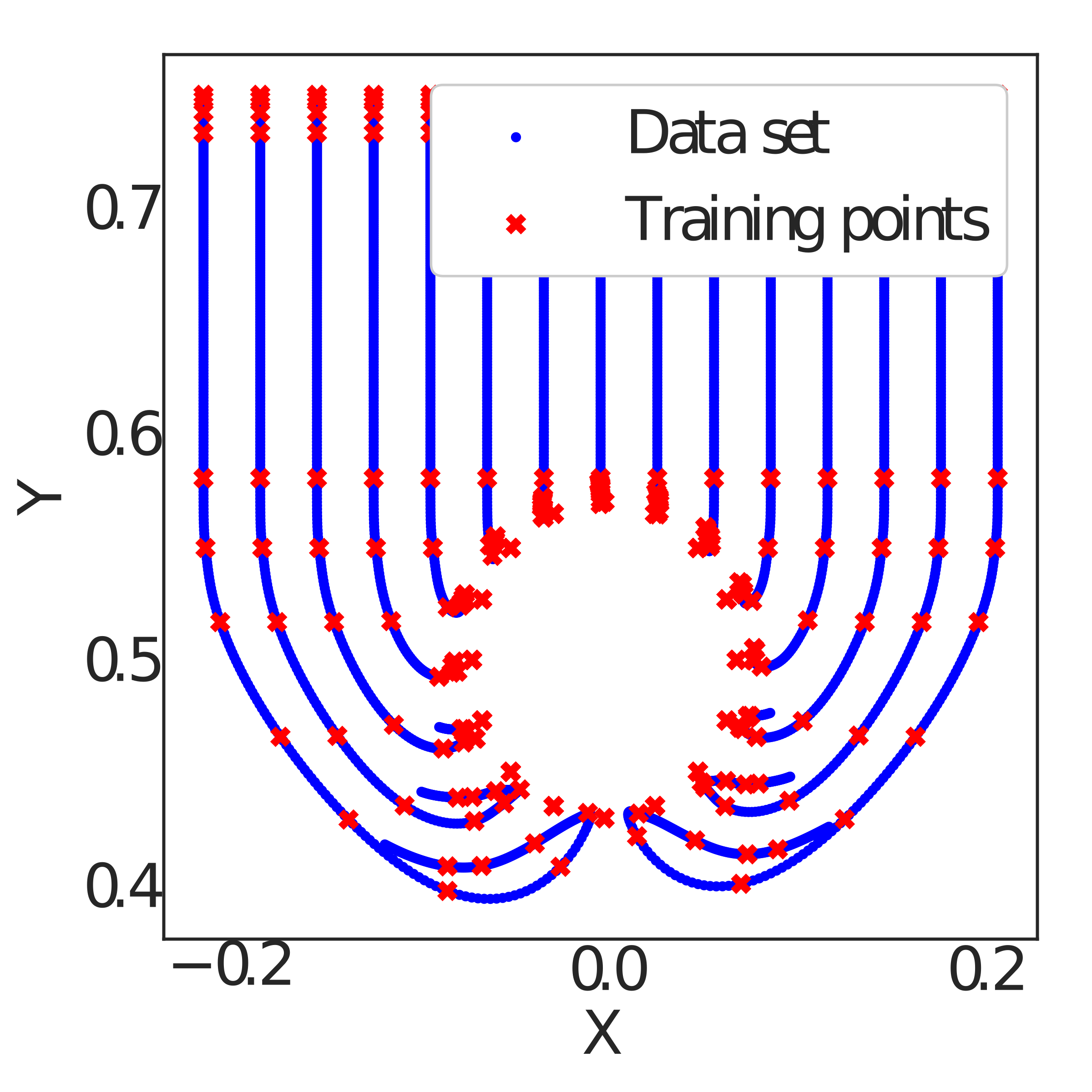}}
	\subfigure[]{\includegraphics[width=0.24\columnwidth]{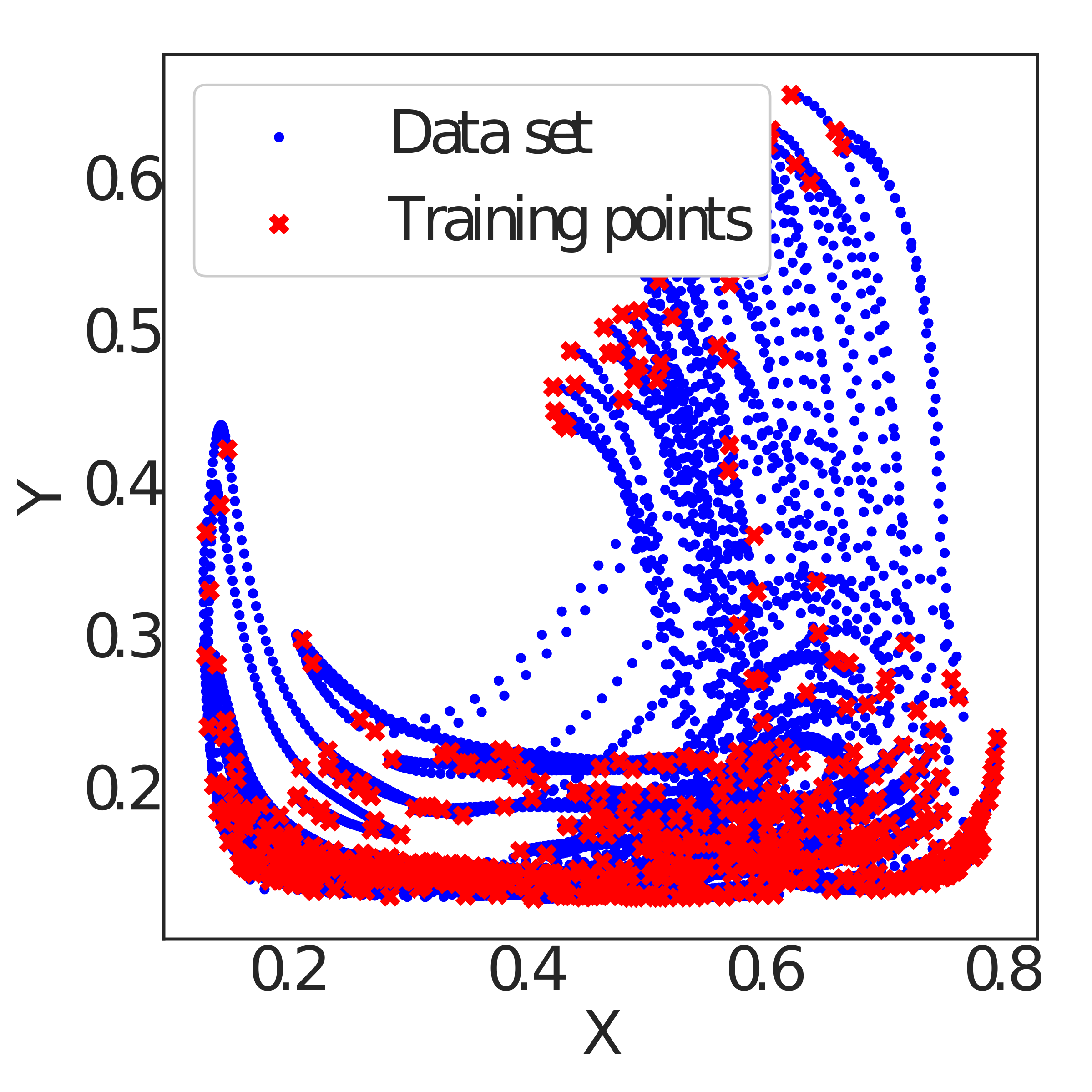}}
	% Targets
	\subfigure[]{\includegraphics[width=0.24\columnwidth]{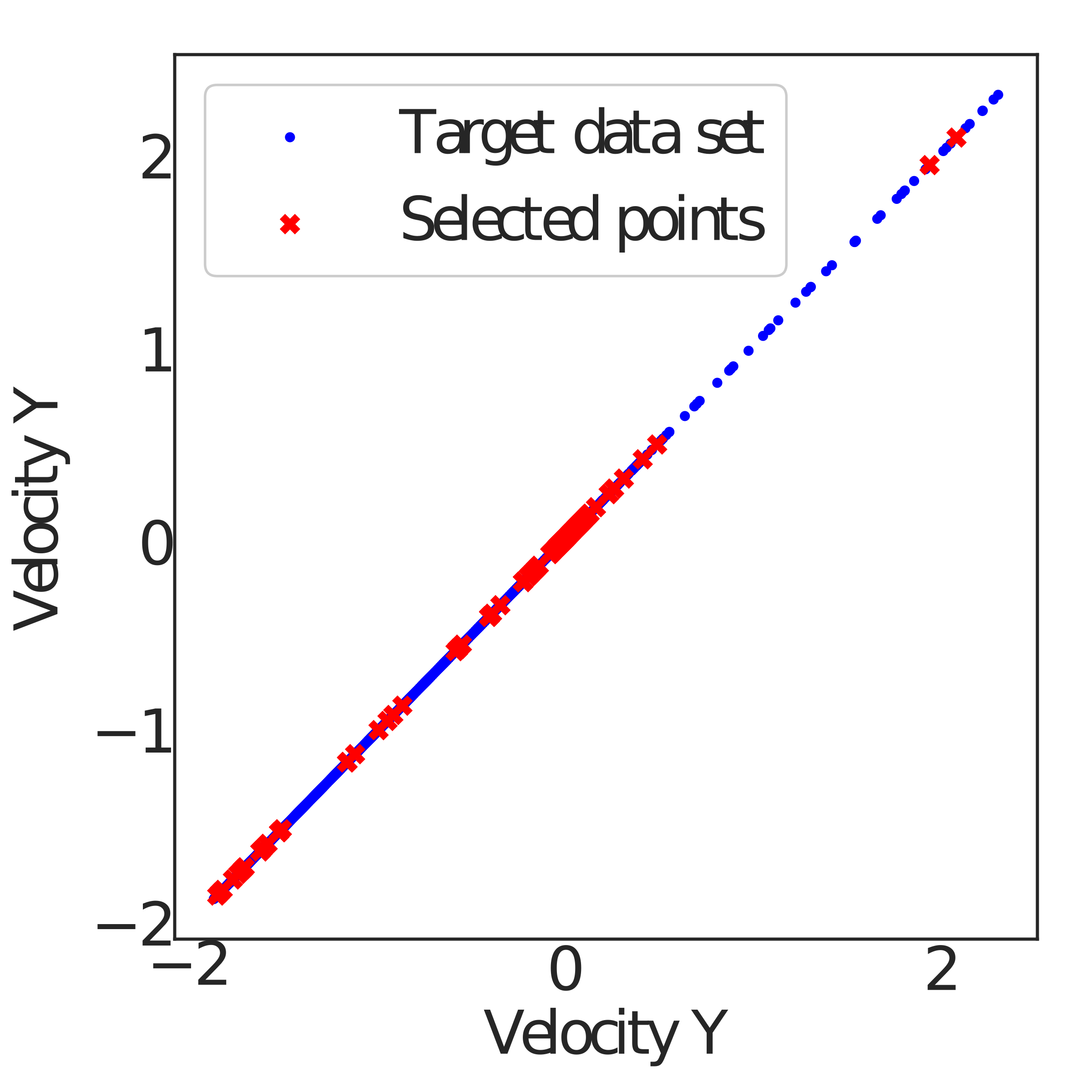}}
	\subfigure[]{\includegraphics[width=0.24\columnwidth]{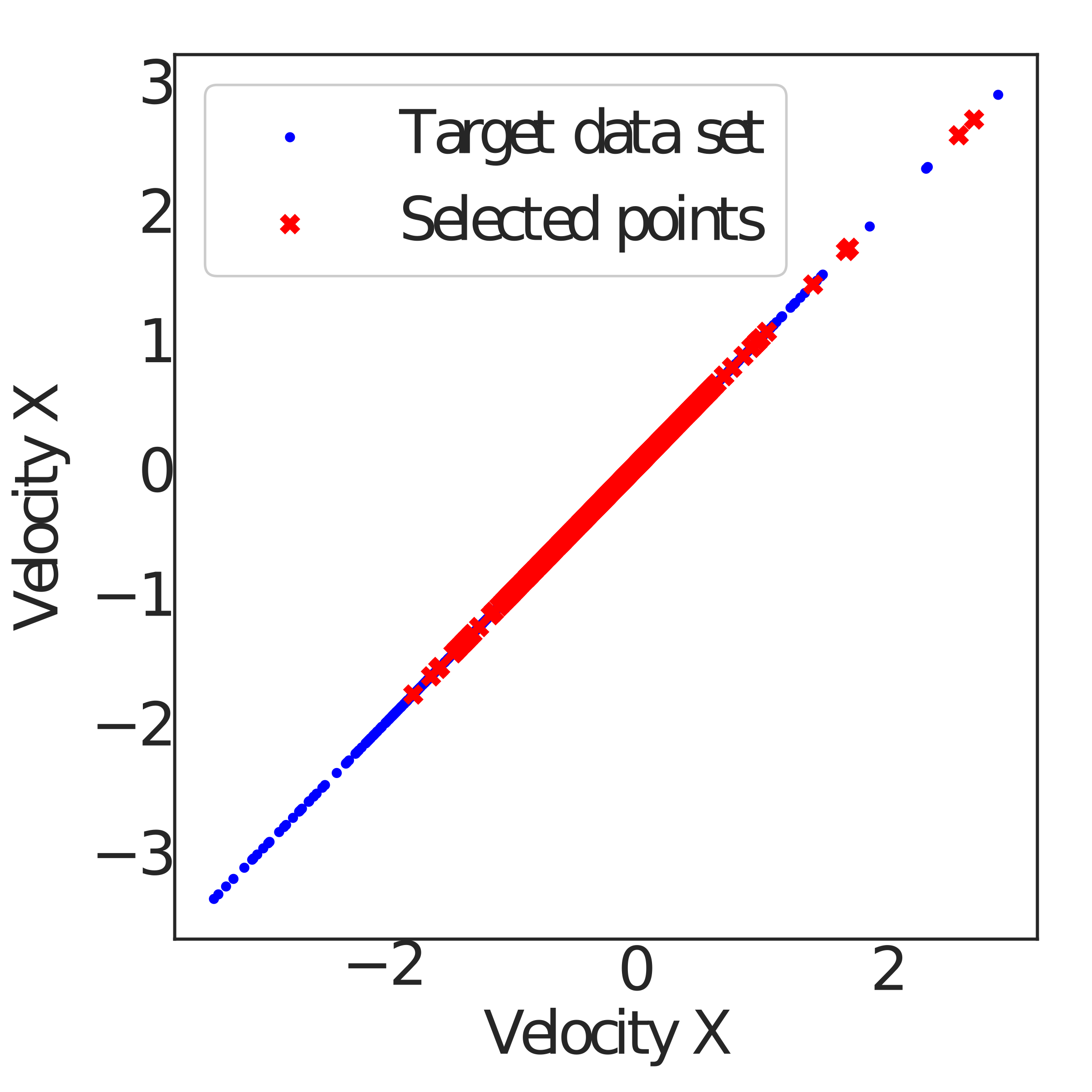}}
	
	\caption{Coordinate state space in (a) GI environment and (b) IEs environment showing full data set (blue) and selected $N=20$ training points (red). Selected training targets (c) $\Delta y_t$ and (d) $\Delta x_t$ in each environment respectively.}
	\label{fig:sup-training-points}
	\vskip 0.1in
\end{figure}

We show the coordinate space for each environment in Figure~\ref{fig:sup-training-points}. The blue markers show the full set of training points whereas the red markers show the limited selected training points, in this case $N=20$. The Figure~\ref{fig:sup-training-points} also shows a target for each of the environments. Note that each node covers a different part of the state-space.

% \newpage

\paragraph{Hardware used for training}
All the experiments were performed in CPU.
For training e-GGP we used a shared server with a CPU Xeon E5-2680 v2.
All the other models were trained in a personal laptop.

\paragraph{Training time}
We provide information about the approximate training time of e-GGP in Table~\ref{tab:supp-training-time} for different number of training points.
We trained e-GGP using Adam with a learning rate of $l_r = 0.1$ for $150$ iterations.
We note that the computational complexity highly depends on the number of nodes in the graph $M$ and the maximum connectivity for a node.
We would like to highlight that our implementation of the kernel used in e-GGP is not ideal and the time taken shown in Table~\ref{tab:supp-training-time} can be highly reduced.

\begin{table}[h]
	\vskip 0.1in
	\caption{Training time of e-GGP for the experiments using a CPU Xeon E5-2680 v2.}
	\label{tab:supp-training-time}
	\begin{center}
		\begin{small}
			\begin{sc}
				\vskip -0.15in
				\begin{tabular}{cccccc}
					\toprule
					& N & $M$ & $W$  & $D$  & $t$  \\
					\toprule
					% \midrule
					\multirow{2}{*}{Graph Interaction}
					& 5 & 31 & 4 & 7 & 14min \\
					& 10 & 31 & 4 & 7 & 1h 20min \\
					& 15 & 31 & 4 &7 & 2h 55min \\
					& 20 & 31 & 4 & 7 & 4h 24min \\
					
					\midrule
					\multirow{4}{*}{Isolated Evolving Sub-graphs}
					& 5 & 44 & 20 & 8 & 33min \\
					& 10 & 44 & 20 & 8 & 2h 16min \\
					& 15 & 44 & 20 & 8 & 5h 42min \\
					& 20 & 44 & 20 & 8 & 14h 7min \\
					\bottomrule
				\end{tabular}
			\end{sc}
		\end{small}
	\end{center}
	\vskip 0.1in
\end{table}

%%%%%% New Appendix section, reset counters
\setcounter{figure}{0}    
\setcounter{table}{0}

\section{Supplementary results}
\label{supp:main-results}

In Section~\ref{sec:experiments}, we provided the results for the Graph Interaction environment.
In the Isolated Evolving Sub-graphs environment, because of the limitation of DGPG to predict test inputs with a different graph structure, we only provided results for a single test set.
Below, we extend those results and detail the results for each test set in the experiments. We also provide an ablation study in the GI environment.

\subsection{e-GGP ablation study}
\label{supp:results-ablation}

\subsection{Graph connectivity ablation study}
\label{sec:exp-evolving-graphs}

We evaluated two variants of our method.
% Explain each variant
The first one assumes that the graph connectivity is fixed (F. e-GGP), using the same adjacency matrix during inference and predictions $A=A_0$.
The second one, has knowledge of the evolving graph structure (E. e-GGP) and uses a different adjacency $A_t$ as the graph evolves over time.

We first compare the results for each of the unseen test sets for $\Delta x_t$ and $\Delta y_t$ in Table~\ref{tab:supp-ablation-x} and Table~\ref{tab:supp-ablation-y} respectively. We can see that the F. e-GGP variant performs slightly better for the offsets close to the training data.
However, the evolving version (E. e-GGP) clearly outperforms the fixed variant shown in Table~\ref{tab:supp-ablation-y}.
This supports our hypothesis and highlights the relevance of accounting for the graph evolution.

\begin{table}[h]
	\vskip 0.1in
	\caption{Results of fixed (F. e-GGP) and evolving (E. e-GGP) variants in 
		GI environment for $\Delta x_t$ with $N=15$.}
	\label{tab:supp-ablation-x}
	\begin{center}
		\begin{small}
			\begin{sc}
				\begin{tabular}{c|cc|cc|cc}
					\toprule
					& \multicolumn{2}{c}{\textbf{RMSE}} &  \multicolumn{2}{c}{\textbf{MAPE}} &  \multicolumn{2}{c}{\textbf{NLL}}\\ 
					\textbf{Offset} & F. e-GGP & E. e-GGP & F. e-GGP & E. e-GGP & F. e-GGP & E. e-GGP \\
					& $\times10^{-2}$ & $\times10^{-2}$ & $\times10^{12}$ & $\times10^{12}$ &  &  \\
					\midrule
					-0.1 & \textbf{0.810} & 0.879 & \textbf{4.14} & 4.52 & \textbf{-57.8} & -54.0 \\
					-0.05 & \textbf{0.890} & 0.956 & \textbf{4.28} & 4.80 & \textbf{-56.7} & -52.7 \\
					0 & \textbf{0.980} & 1.04 & \textbf{3.89} & 3.92 & \textbf{-54.8} & -50.4 \\
					0.05 & \textbf{1.10} & 1.14 & 4.57 & \textbf{4.17} & \textbf{-52.5} & -48.3 \\
					0.1 & \textbf{1.40} & 1.41 & 10.8 & \textbf{9.80} & \textbf{-47.1} & -41.4 \\
					0.2 & \textbf{1.56} & \textbf{1.56} & 16.1 & \textbf{15.4} & \textbf{-44.7} & -38.3 \\
					0.3 & \textbf{1.87} & \textbf{1.87} & 29.5 & \textbf{29.4} & \textbf{-43.2} & -37.5 \\
					\bottomrule
				\end{tabular}
			\end{sc}
		\end{small}
	\end{center}
	% \vskip 0.1in
\end{table}

\begin{table}[h]
	\vskip 0.1in
	\caption{Results of fixed (F. e-GGP) and evolving (E. e-GGP) variants in 
		GI environment for $\Delta y_t$ with $N=15$.}
	\label{tab:supp-ablation-y}
	\begin{center}
		\begin{small}
			\begin{sc}
				\begin{tabular}{c|cc|cc|cc}
					\toprule
					& \multicolumn{2}{c}{\textbf{RMSE}} &  \multicolumn{2}{c}{\textbf{MAPE}} &  \multicolumn{2}{c}{\textbf{NLL}}\\ 
					\textbf{Offset} & F. e-GGP & E. e-GGP & F. e-GGP & E. e-GGP & F. e-GGP & E. e-GGP \\
					& $\times10^{-2}$ & $\times10^{-2}$ & $\times10^{-1}$ & $\times10^{-1}$ &  &  \\
					\midrule
					-0.1 & 2.52 & \textbf{2.30} & 3.98 & \textbf{3.79} & -13.91 & \textbf{-18.14} \\
					-0.05 & 2.42 & \textbf{2.36} & 3.98 & \textbf{3.79} & \textbf{-17.98} & -16.00 \\
					0 & \textbf{2.17} & 2.31 & 3.32 & \textbf{2.86} & \textbf{-31.15} & -20.89 \\
					0.05 & \textbf{2.25} & 2.45 & 1.74 & \textbf{1.73} & \textbf{-33.17} & -23.30 \\
					0.1 & 3.61 & \textbf{3.31} & \textbf{1.04} & \textbf{1.04} & \textbf{-18.80} & -17.10 \\
					0.2 & 4.63 & \textbf{3.98} & 5.01 & \textbf{4.69} & -11.09 & \textbf{-12.10} \\
					0.3 & 6.84 & \textbf{5.46} & 0.861 & \textbf{0.732} & 0.20 & \textbf{-3.97} \\
					\bottomrule
				\end{tabular}
			\end{sc}
		\end{small}
	\end{center}
	\vskip 0.1in
\end{table}

\newpage

\subsection{Results on scarce data for the graph interaction environment}
\label{supp:results-rope}
\begin{figure}[h]
	\vskip 0.1in
	\centering
	\subfigure[]{\label{fig:rmse-a}\includegraphics[width=0.325\columnwidth]{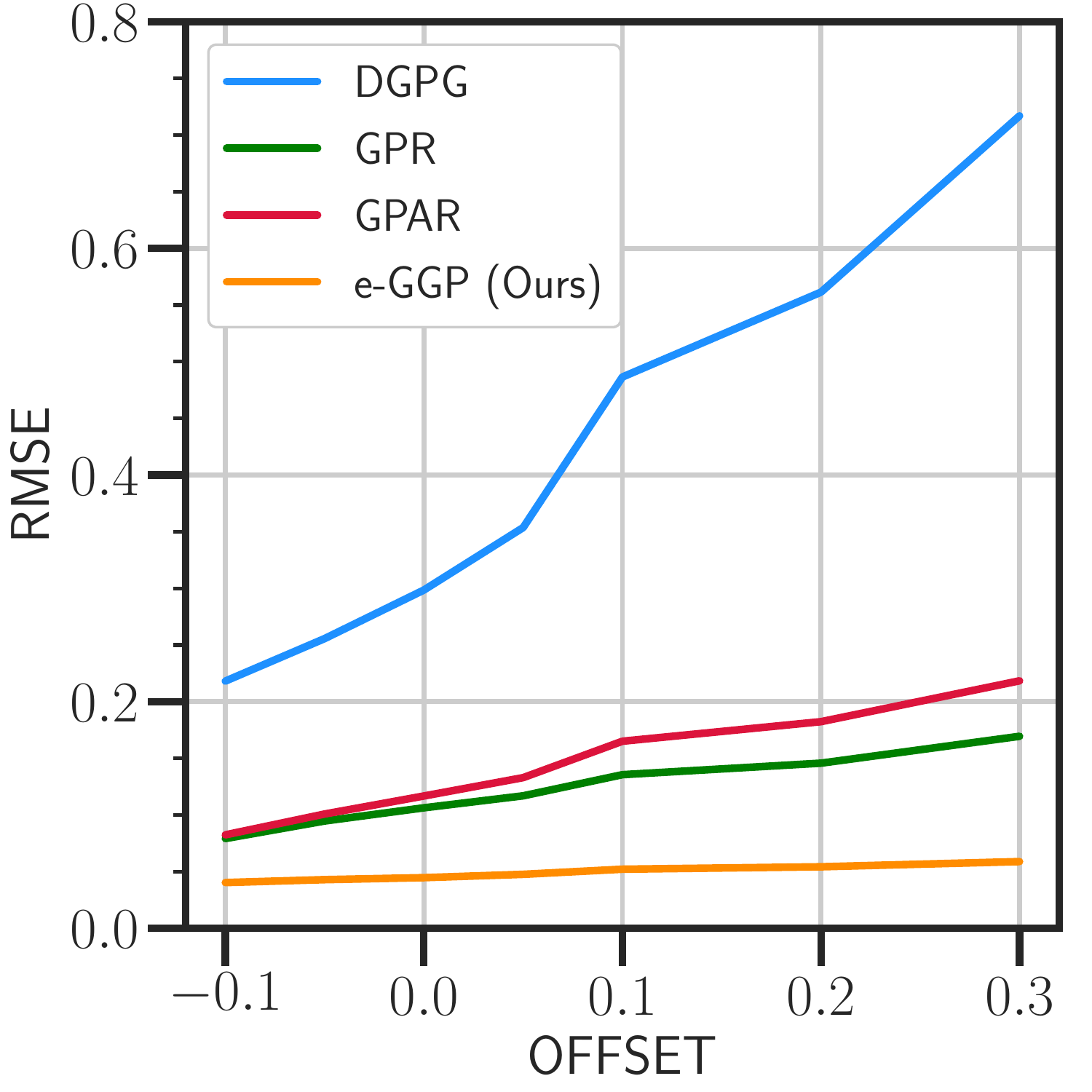}}
	\subfigure[]{\label{fig:rmse-b}\includegraphics[width=0.325\columnwidth]{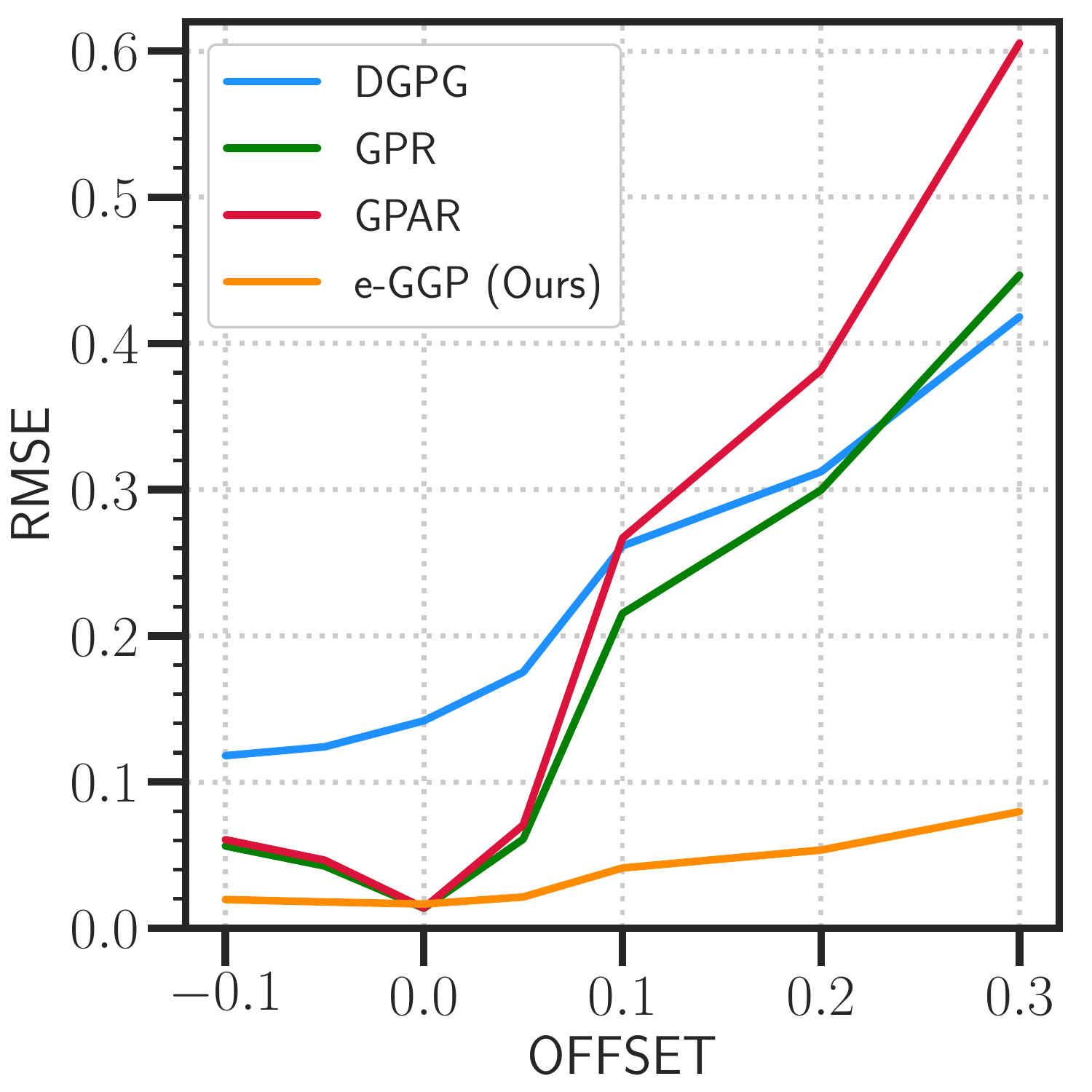}}
	\caption{Comparison of one-step ahead prediction of RMSE results for GPR, GPAR, DGPG and e-GGP in the graph interaction experiment for (a) 10 training points and (b) 20 training points.}
	\label{fig:supp-exp_results_rope_rmse}
	\vskip 0.1in
\end{figure}

In this section, we provide additional results for the GI environment. We show in Figure~\ref{fig:supp-exp_results_rope_rmse} the RMSE results for different offsets of the rope. The rope offset used for training was $Z=0$. We can see that e-GGP outperforms the other methods for both $N=10$ and $N=20$ training points.

\subsection{Results on scarce data for the isolated evolving sub-graphs environment}
\label{supp:results-water}

We provide the comparison of e-GGP and the baselines GPR, GPAR for the node velocity prediction for each target $\Delta x_t$ and $\Delta y_t$ in Table~\ref{tab:supp-results-water-x} and Table~\ref{tab:supp-results-water-y} respectively. 
We discussed in Section~\ref{sec:exp-scarce-data} the averaged results of e-GGP. In here, we can see that for both targets e-GGP has consistent low RMSE, MAPE and NLL.  These results show that our model is capable of learning the transitions more accurately, regardless of the target.

\begin{table}[h]
	\vskip 0.15in
	\caption{Results for GPR, GPAR and e-GGP in evolving isolated sub-graphs test datasets for the target $\Delta x_t$.}
	\label{tab:supp-results-water-x}
	\begin{center}
		\begin{small}
			\begin{sc}
				\vskip 0.15in
				\begin{tabular}{cc|ccc|ccc|ccc}
					\toprule
					% \midrule
					&  & \multicolumn{3}{c}{\textbf{RMSE}} &  \multicolumn{3}{c}{\textbf{MAPE}} &  \multicolumn{3}{c}{\textbf{NLL}}\\
					N & $M$ & \textbf{GPR} & \textbf{GPAR} & \textbf{e-GGP} & \textbf{GPR} & \textbf{GPAR} & \textbf{e-GGP} & \textbf{GPR} & \textbf{GPAR} & \textbf{e-GGP} \\
					\midrule
					
					\multirow{2}{*}{5} &
					55 & 0.162 & 0.194 & \textbf{0.107} & 1.27 & 2.06 & \textbf{1.09} & \textbf{-40.173} & 242.025 & 3.542 \\
					& 66 & 0.306 & 0.307 & \textbf{0.148} & 3.03 & 4.95 & \textbf{2.23} & \textbf{-22.676} & 435.562 & 89.792 \\
					\midrule
					\multirow{2}{*}{10} &
					55 & 0.148 & 0.172 & \textbf{0.110} & 0.97 & 1.32 & \textbf{0.88} & \textbf{-43.425} & 83.373 & 3.574 \\
					& 66 & 0.397 & 0.277 & \textbf{0.154} & \textbf{2.33} & 3.31 & 2.37 & \textbf{-18.255} & 370.486 & 193.102 \\
					\midrule
					\multirow{2}{*}{15} &
					55 & 0.133 & 0.126 & \textbf{0.096} & 0.86 & 0.86 & \textbf{0.79} & -49.657 & 146.950 & \textbf{-90.342} \\
					& 66 & 0.325 & 0.295 & \textbf{0.152} & 2.41 & 2.56 & \textbf{2.16} & -30.424 & 490.559 & \textbf{-74.796} \\
					\midrule
					\multirow{2}{*}{20} &
					55 & 0.117 & 0.111 & \textbf{0.086} & 0.78 & \textbf{0.65} & \underline{0.70} & -49.822 & 5.2$\times 10^{6}$ & \textbf{-96.042} \\
					& 66 & 0.334 & 0.309 & \textbf{0.194} & 2.04 & \textbf{1.76} & 2.10 & -30.945 & 1.07$\times 10^{7}$ & 1.98$\times 10^{6}$ \\
					
					\bottomrule
				\end{tabular}
			\end{sc}
		\end{small}
	\end{center}
	\vskip -0.1in
\end{table}

\newpage

\begin{table}[H]
	\vskip 0.15in
	\caption{Results for GPR, GPAR and e-GGP in evolving isolated sub-graphs test datasets for the target $\Delta y_t$.}
	\label{tab:supp-results-water-y}
	\begin{center}
		\begin{small}
			\begin{sc}
				\begin{tabular}{cc|ccc|ccc|ccc}
					\toprule
					% \midrule
					&  & \multicolumn{3}{c}{\textbf{RMSE}} &  \multicolumn{3}{c}{\textbf{MAPE}} &  \multicolumn{3}{c}{\textbf{NLL}}\\
					N & $M$ & \textbf{GPR} & \textbf{GPAR} & \textbf{e-GGP} & \textbf{GPR} & \textbf{GPAR} & \textbf{e-GGP} & \textbf{GPR} & \textbf{GPAR} & \textbf{e-GGP} \\
					\midrule
					
					\multirow{2}{*}{5} &
					55 & 0.403 & 0.396 & \textbf{0.264} & \textbf{2.52} & 3.25 & 3.39 & \textbf{-21.474} & 92.63 & \underline{-0.256} \\
					& 66 & 0.643 & 0.751 & \textbf{0.567} & 4.48 & \textbf{3.70} & 6.07 & \textbf{5.156} & 167.1 & 200.3 \\
					\midrule
					\multirow{2}{*}{10} &
					55 & 0.414 & 0.429 & \textbf{0.318} & 1.38 & 1.16 & \textbf{1.07} & -47.420 & \textbf{-114.1} & \underline{-110.4} \\
					& 66 & \textbf{0.647} & 0.654 & 0.713 & 3.54 & 8.73 & \textbf{1.39} & -13.558 & \textbf{-55.81} & -26.29 \\
					\midrule
					\multirow{2}{*}{15} &
					55 & 0.435 & \textbf{0.391} & 0.418 & 1.07 & \textbf{0.93} & 1.48 & -42.35 & \textbf{-111.38} & -50.755 \\
					& 66 & 0.645 & 0.632 & \textbf{0.596} & 4.87 & \textbf{1.45} & \underline{1.83} & -9.78 & \textbf{-44.65} & 95.468 \\
					\midrule
					\multirow{2}{*}{20} &
					55 & 0.383 & 0.407 & \textbf{0.253} & 1.00 & \textbf{0.78} & 1.09 & 2.39 $\times 10^{5}$ &  \textbf{-50.321} & 1.17 $\times 10^{6}$ \\
					& 66 & 0.596 & 0.668 & \textbf{0.435} & 3.04 & \textbf{0.96} & 2.19 &  3.65 $\times 10^{5}$ & \textbf{-19.752} & 2.47 $\times 10^{6}$ \\
					\bottomrule
				\end{tabular}
			\end{sc}
		\end{small}
	\end{center}
	\vskip -0.1in
\end{table}

\end{document}